\documentclass[
  preprint,
  twocolumn,
  5p,  
  number,  % numbered citation style
  sort&compress,  % sort and compress citations
]{elsarticle}
\journal{ISPRS Journal of Photogrammetry and Remote Sensing}
\def\github{\url{https://github.com/jnyborg/timematch}}

\usepackage{hyperref}
\usepackage{url}
\usepackage{amsmath}
\usepackage{amssymb}

\usepackage{bm}      % for \bm macro
\usepackage{bbm}  % for indicator function
\usepackage{booktabs}
\usepackage{microtype}
\usepackage{graphicx}  % includegraphics
\usepackage{subcaption}  % subfigures
\usepackage{mleftright}  % mleft and mright for functions
\usepackage{xcolor}
\usepackage{array}

\usepackage{tabu}  % row color command - remove after revision

\usepackage{pifont}
\newcommand{\cmark}{\color[rgb]{0,.5,0}\ding{51}}%
\newcommand{\xmark}{\color[rgb]{.5,0,0}\ding{55}}%

\newcommand{\figfile}[1]{pdf/#1.pdf}

\usepackage[ruled,vlined,linesnumbered]{algorithm2e}
    \SetArgSty{textnormal}  % disable italic text

\newcommand{\R}{\mathbb{R}}
\newcommand{\D}{\mathcal{D}}
\newcommand{\E}{\mathbb{E}}
\DeclareMathOperator*{\argmax}{argmax} % thin space, limits underneath in displays
\DeclareMathOperator*{\argmin}{argmin} % thin space, limits underneath in displays

\newcommand{\blue}[1]{#1}
\newcommand{\green}[1]{#1}

\makeatletter
\DeclareRobustCommand\onedot{\futurelet\@let@token\@onedot}
\def\@onedot{\ifx\@let@token.\else.\null\fi\xspace}

\def\eg{\emph{e.g}\onedot} 
\def\ie{\emph{i.e}\onedot}

\def\etal{\emph{et al}\onedot}
\makeatother

\usepackage{mathtools}
\DeclarePairedDelimiterX{\infdivx}[2]{(}{)}{%
  #1\;\delimsize\|\;#2%
}
\newcommand{\kld}{D_{\mathrm{KL}}\infdivx}

\begin{document}
\begin{frontmatter}

\title{TimeMatch: Unsupervised Cross-Region Adaptation by Temporal Shift Estimation}

\author[1,3]{Joachim Nyborg}
\author[2]{Charlotte Pelletier}
\author[2]{Sébastien Lefèvre}
\author[1]{Ira Assent}
\address[1]{Department of Computer Science, Aarhus University, Aarhus, Denmark}
\address[2]{IRISA UMR 6074, Université Bretagne Sud, Vannes, France}
\address[3]{FieldSense A/S, Aarhus, Denmark}

\begin{abstract}
The recent developments of deep learning models that capture complex temporal patterns of crop phenology have greatly advanced crop classification from Satellite Image Time Series (SITS). However, when applied to target regions spatially different from the training region, these models perform poorly without any target labels due to the temporal shift of crop phenology between regions. 
\blue{Although various unsupervised domain adaptation techniques have been proposed in recent years, no method explicitly learns the temporal shift of SITS and thus provides only limited benefits for crop classification.}
To address this, we propose TimeMatch, which explicitly accounts for the temporal shift for improved SITS-based domain adaptation.
\blue{In TimeMatch, we first estimate the temporal shift from the target to the source region using the predictions of a source-trained model. Then, we re-train the model for the target region by an iterative algorithm where the estimated shift is used to generate accurate target pseudo-labels.}
Additionally, we introduce an open-access dataset for cross-region adaptation from SITS in four different regions in Europe. On our dataset, we demonstrate that TimeMatch outperforms all competing methods by 11\% in average F1-score across five different adaptation scenarios, setting a new state-of-the-art in cross-region adaptation. \green{Our source code and dataset are available at~\github}. 
\end{abstract}

\end{frontmatter}

\section{Introduction}
\label{sec:introduction}
Today, the availability of satellite image time series (SITS) data is rapidly increasing.
For instance, the twin Sentinel-2 satellites provide imagery of the entire Earth every two to five days~\cite{drusch2012sentinel}.
A frequent acquisition of images is crucial for vegetation-related remote sensing applications such as crop type classification~ \cite{reed1994measuring, vuolo2018much}. Multi-temporal data enables capturing the phenological development of crops (\ie, the progressions of crop growth), a key dimension to discriminate each crop type~\cite{odenweller1984crop}.
Recently, the increasing availability of SITS along with advances in deep learning has led to crop classifiers with temporal neural architectures using convolutions~\cite{pelletier2019temporal, zhong2019deep}, recurrent units~\cite{ndikumana2018deep, russwurm2017temporal, ienco2017land, minh2018deep}, self-attention~\cite{russwurm2020self, garnot2020satellite}, or combinations thereof~\cite{russwurm2018multi, interdonato2019duplo}. 

\begin{figure}[ht]
\centering
\includegraphics[width=1.0\linewidth]{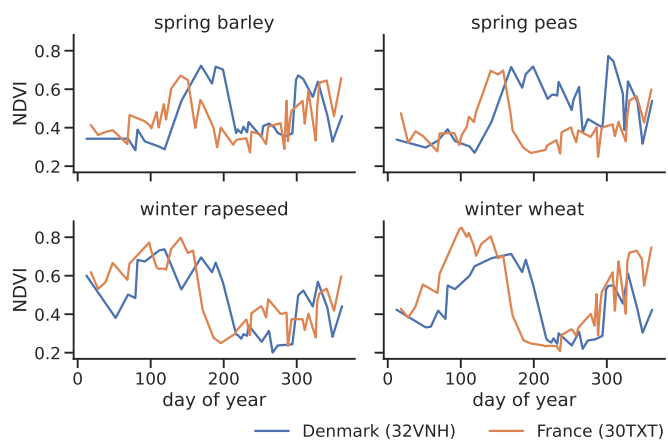}
\caption{
Normalized difference vegetation index (NDVI) time series for crops from two different Sentinel-2 tiles in Europe, indicating the growth of four crop types. Crops develop similarly in different regions, but the patterns are temporally shifted, \eg if crops ripen at different times of the year.}%, causing crop classifiers trained in one region to fail when applied to another.}
\label{fig:ndvi}
\end{figure}

These crop classification models achieve impressive performance by capturing the temporal structure of the problem but rely on the existence of a large amount of labeled training data. While unlabeled SITS are plenty, access to labels in the region of interest (the \emph{target} domain) is often either costly or otherwise unavailable. A possible solution is to train a model in a region with labels available (the \emph{source} domain) and apply it to the unlabeled target region. However, when the two regions are geographically different, the dissimilarity between the source and target data distributions can cause a source-trained model to perform poorly when applied to the target region~\cite{tuia2016domain, lucas2021bayesian, kondmann2021denethor}. 

Addressing the distributional shift problem to adapt a source-trained model to an unlabeled target domain is in machine learning known as unsupervised domain adaptation (UDA)~\cite{pan2009survey, tuia2016domain, kellenberger2021deep}. 
Here, we consider the cross-region UDA problem for SITS~\cite{wang2021phenology}, where we are provided with labeled data from a source region and unlabeled data from a target region. In this setting, the source and target data distributions differ due to changes in local conditions, such as the soil, climate, and farmer practices, which cause spectral and temporal shifts~\cite{tuia2016domain}.

Addressing the temporal shift is of particular importance when adapting crop classifiers to new regions, as we illustrate in Figure~\ref{fig:ndvi}.
While crops in different regions have similar growth patterns, the timing of key growth stages, such as the peak of greenness, is shifted along the temporal axis. As crops are classified primarily by their unique growth patterns, the temporal shift may cause misclassifications when a source-trained model is applied to a target region. For example, the shift in time could cause the phenology of spring barley to appear similar to that of winter barley in the target. Thus, accounting for the temporal shift is a key factor in cross-region adaptation.

\blue{A possible approach could be to train models that are invariant to temporal shifts, such as by applying random temporal shifts to the training data. 
However, as the temporal shift could be the main feature that separates two crop types, 
shift-invariant models have reduced classification ability compared to shift-variant models.}

Another approach is to apply existing UDA methods.
Typically, these methods address domain adaptation by constraining the classifier to operate on domain-invariant features~\cite{ben2010theory}. This is achieved by training the classifier to perform well on the source domain while minimizing a divergence measure between features extracted from the source and the target domains~\cite{ganin2015unsupervised, tzeng2014deep, wang2021phenology}. While these methods have been successfully applied in various applications~\cite{wilson2020survey, kellenberger2021deep}, they do not directly account for the temporal shift in SITS and have thus been reported to provide limited benefits in cross-region UDA~\cite{lucas2020unsupervised}. 
\blue{More recently, self-training methods have emerged as a promising alternative to domain-invariant methods~\cite{saito2017asymmetric, shu2018dirt, morerio2020generative, chen2020adversarial, zou2019confidence}.
Self-training iteratively generates pseudo-labels~\cite{lee2013pseudo} for the target domain and then uses them to retrain the model with target data. To account for noisy pseudo-labels caused by the domain shift, these methods typically incorporate a refinement step where the noise is reduced in various ways, such as with generative models~\cite{morerio2020generative} or learned confusion matrices~\cite{chen2020adversarial}. Still, no method considers the particular case of SITS where the pseudo-label noise is caused by a temporal shift.}

\blue{In this paper, we propose \emph{TimeMatch}, a self-training method for cross-region UDA where we directly account for the temporal shift.
TimeMatch consists of two components: (i) the temporal shift estimation and (ii) the TimeMatch learning algorithm.}

\blue{Estimating the temporal shift directly from the target data is difficult, as the lack of labels hinders \eg the comparison of class-wise vegetation indices as in Figure~\ref{fig:ndvi}. To address this, we propose an unsupervised method where we estimate the temporal shift 
from target to source with a source-trained model. First, we obtain the softmax predictions of the model when input target data with different temporal shifts. Then, we choose the temporal shift with high prediction confidences across a diverse set of classes. 
We show that this approach corresponds well to the actual climatic differences between the two regions.
Moreover, as correctly classified examples tend to have higher prediction confidence~\cite{hendrycks2016baseline}, the estimated shift enables us to generate more accurate pseudo-labels in the target domain for self-training.}

\blue{In TimeMatch learning, we therefore use self-training to adapt a model to the target domain. 
We propose an iterative algorithm where we alternate between temporal shift estimation and re-training the model for the target domain by learning from both labeled source data and pseudo-labeled target data. By doing so, the model learns discriminative target features for accurate crop classification in the target region.}

Lastly, we present the TimeMatch dataset, a challenging new open-access dataset for training and evaluating cross-region models on SITS with over 300.000 annotated parcels from four different regions in Europe.
Evaluated on this dataset, our approach outperforms all competing methods by 11\% F1-score on average  across five different cross-region UDA experiments.

In summary, our contributions are as follows:
\begin{itemize}
    \item We propose a method for estimating the temporal shift between a labeled source region and an unlabeled target region to reduce their temporal discrepancy.
    \item We propose \emph{TimeMatch}, a novel UDA method designed for the cross-region problem of SITS, where crop classification models are adapted to an unlabeled target region by \blue{self-training} on temporally shifted data for improved performance compared to existing methods. \green{Our source code is available at~\github}.
    \item We release the TimeMatch dataset~\cite{nyborg2021timematchdataset}, a new dataset for training and evaluating cross-region UDA models on SITS from four different European regions.
\end{itemize}

This paper is organized as follows. Section~\ref{sec:relatedwork} describes the existing literature related to our work. Section~\ref{sec:method} describes the proposed method for temporal shift estimation and the TimeMatch learning algorithm. Section~\ref{sec:data_and_material} presents our dataset and the experimental setup, and Section~\ref{sec:experiments} the experimental results. Lastly, Section~\ref{sec:conclusion} concludes this work.

\section{Related Work}
\label{sec:relatedwork}
TimeMatch is related to the existing work in unsupervised domain adaptation of learning domain-invariant features, time-series adaptation, cross-region adaptation, and self-training.

\subsection{Domain-Invariant Methods}
The predominant approach in UDA is to train the classifier to rely only on domain-invariant features~\cite{ben2010theory, wilson2020survey}.
To this end, several works consider adversarial training~\cite{ganin2015unsupervised, ganin2016domain, long2017conditional}. In domain adversarial neural networks (DANN)~\cite{ganin2015unsupervised, ganin2016domain}, the feature extractor is adversarially trained to produce domain-invariant features that are indistinguishable by a domain discriminator. Conditional domain adversarial networks (CDAN)~\cite{long2017conditional}  improves upon DANN by conditioning the domain discriminator on classifier predictions in addition to features to enable the alignment of multimodal data distributions. 

Another approach is to align the feature distributions directly by minimizing a divergence measure. Choices for divergence measure include maximum mean discrepancy (MMD)~\cite{tzeng2014deep}, correlation alignment~\cite{sun2016deep}, or optimal transport~\cite{damodaran2018deepjdot, fatras2021jumbot}. Recently, JUMBOT~\cite{fatras2021jumbot} achieves state-of-the-art UDA results by using mini-batch unbalanced optimal transport to minimize the domain discrepancy of joint deep feature and label distributions.

\blue{While domain-invariant methods achieve strong results on computer vision datasets, they do not explicitly handle the temporal dimensions of SITS data and time series in general.}
 
\subsection{Time-Series Unsupervised Domain Adaptation}
Few methods tackle the challenge of time series UDA. Current methods for time series are typically also based on learning domain-invariant features~\cite{purushotham2016vrada, wilson2020codats, bailly2017nonlinear}.
Recurrent domain adversarial neural network (R-DANN) and variational recurrent adversarial deep domain adaptation (VRADA) explore long short-term memory and variational recurrent neural networks as feature extractors, respectively, and learn domain-invariant features using the DANN method~\cite{purushotham2016vrada}. Likewise, the convolutional deep domain adaptation model for time series data (CoDATS) learns domain-invariant features with a temporal convolutional network with the DANN method~\cite{wilson2020codats}.
However, while these methods are effective at learning domain-invariant features for time series, 
they are not designed to learn the temporal shift present in SITS.

\subsection{Cross-Region Crop Classification}
Lucas~\etal~\cite{lucas2020unsupervised} reports that existing UDA methods, including existing domain-invariant methods~\cite{gong2012geodesic, fernando2013unsupervised}, perform poorly when applied to cross-region UDA of SITS due to the temporal shift problem and the change in class distribution between the two regions. 
Recently, Wang~\etal~\cite{wang2021phenology} proposed the phenology alignment network (PAN) as the first method for cross-region UDA of SITS. PAN learns domain-invariant features with MMD~\cite{tzeng2014deep} and a feature extractor consisting of gated recurrent units and self-attention. Still, as PAN learns domain-invariant features, the temporal shift problem is not directly addressed. 

\subsection{Self-Training Methods}
\blue{
Semi-supervised learning (SSL) is a similar task to domain adaptation, but where the labeled and unlabeled data are sampled from the same data distribution~\cite{chapelle2009semi}. Many SSL methods are based on pseudo-labeling~\cite{lee2013pseudo} (also called self-training~\cite{sohn2020fixmatch}), where the model's own high-confidence predictions are used as labels for the unlabeled samples.
In Mean Teacher~\cite{tarvainen2017mean}, the model assumes a dual role as \emph{teacher} and \emph{student}. 
The student is updated by gradient descent with pseudo-labels generated by the teacher, whereas the teacher is updated by an exponential moving average (EMA) of student parameters to reduce pseudo-label noise.
FixMatch~\cite{sohn2020fixmatch} generates pseudo-labels for weakly-augmented inputs, and uses confident pseudo-labels to self-train the model on strongly-augmented inputs, regularizing the model to output consistent pseudo-labels for random augmentations of the input.
}

\blue{Recently, self-training has emerged for UDA as an alternative to domain-invariant methods~\cite{saito2017asymmetric, shu2018dirt, morerio2020generative, chen2020adversarial, zou2019confidence}.
By learning from both labeled source data and pseudo-labeled target data, self-training methods implicitly encourage feature alignment for each class without restricting the model to operate on domain-invariant features. However, since the domain shift often results in increased pseudo-label noise compared to SSL, existing methods introduce various refinement methods to reduce the noise, such as co-training~\cite{chen2011co}, tri-training~\cite{saito2017asymmetric}, conditional generative models~\cite{morerio2020generative}, or confidence regularization~\cite{zou2019confidence}.
Recently, Adversarial-Learned Loss for Domain Adaptation (ALDA)~\cite{chen2020adversarial} proposes to refine the pseudo-labels with a noise-correcting domain discriminator.}

\blue{Similar to this line of work, our approach is based on self-training. 
By directly accounting for the temporal shift, we can temporally align the target SITS with that of the source, which enables the generation of more accurate pseudo-labels compared to existing self-training methods that do not.}

\begin{figure*}[ht]
\centering
\includegraphics[width=0.8\linewidth]{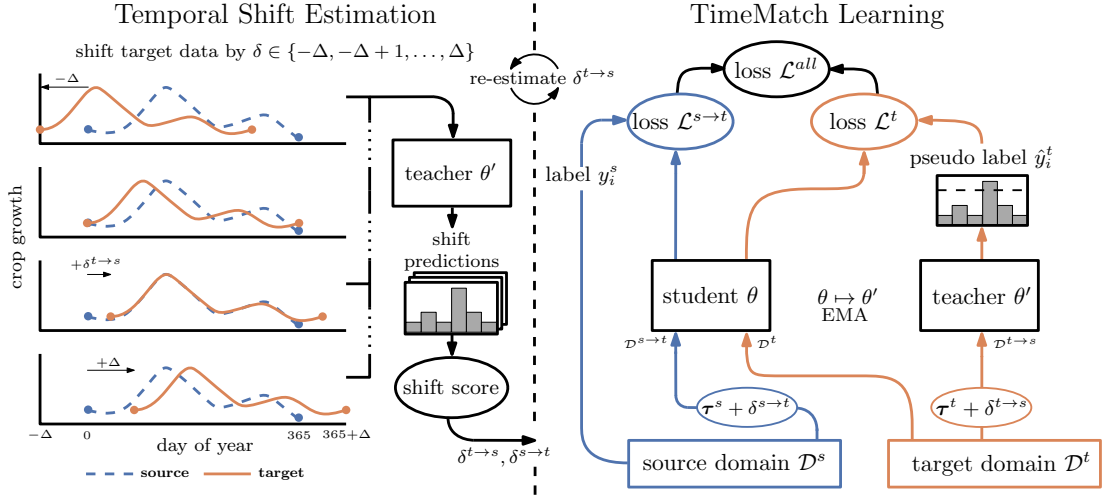}
\caption{Overview of TimeMatch. Both the student and teacher are pre-trained on the source domain. \textit{Temporal Shift Estimation}: {
        We input shifted target data to the teacher model and obtain its predictions for each shift. We then score each shift by the confidence and diversity of the teacher predictions, and the shift with the best score is output as the temporal shift estimate $\delta^{t \to s}$ and $\delta^{s \to t} = -\delta^{t \to s}$.
    }
    \textit{TimeMatch Learning}:{
        The teacher generates pseudo-labels for unlabeled target data shifted by $\delta^{t \to s}$. 
        Then, the student is updated for (non-shifted) target data using the pseudo-labels, and for source data shifted by 
        $\delta^{s \to t}$ using the available source labels. 
        As a result, the student is adapted to the target domain with both generated target labels and actual source labels. 
        After the student parameters have been updated with gradient descent, the teacher parameters are updated as an exponential moving average (EMA) of the student parameters. 
        As both models adapt to the temporal shift of the target domain, the best shift for pseudo-labeling with the teacher changes and must be re-estimated.
        The EMA ensures the teacher adapts slowly which enables $\delta^{t\to s}$ to be re-estimated each epoch only for improved training efficiency and pseudo-label accuracy.
    }
}
\label{fig:methods_figure}
\end{figure*}

\section{TimeMatch}
\label{sec:method}
In this section, we describe our proposed method TimeMatch for cross-region UDA. We begin by formally defining the problem setting, followed by an overview of how TimeMatch addresses it. We then give the details of the two TimeMatch components: temporal shift estimation and TimeMatch learning.

\subsection{Problem Setting}
\label{sec:problem}
In crop classification, the input is a sequence of satellite images $\bm x_i = (\bm x_i^{(1)}, \dots, \bm x_i^{(T_i)})$ of length $T_i$ 
to be classified into one of the $K$ crop classes. In object-based classification, which we focus on in this work, each $\bm x_i \in \R^{T_i \times N_i \times C}$ contains a sequence of $N_i$  pixels of $C$ spectral bands \blue{within a homogeneous, agricultural plot of land which we refer to as a \textit{parcel}.}

Each $\bm x_i$ is accompanied by a sequence $\bm \tau_i = (\tau_i^{(1)}, \dots, \tau_i^{(T_i)})$ indicating the time $\tau_i^{(j)}$ at which each observation $\bm x_i^{(j)}$ is sampled. In practice, $\tau_i^{(j)}$ is typically represented by the \blue{day-of-year}~\cite{russwurm2017temporal, garnot2020satellite}, and makes it possible to account for the irregular temporal sampling of most satellites. 
The goal of the crop classification task is to learn a model which predicts class probabilities $p(y | (\bm x_i, \bm \tau_i)) \in \R^K$, typically learned with supervision from labels $y \in \{1, \dots, K\}$.

In this work, we consider the problem of cross-region UDA. We are given a source domain $\D^s = \{(\bm x_i^s, \bm \tau_i^s, y_i^s)\}_{i=1}^{n^s}$ of $n^s$ labeled SITS and a target domain $\D^t = \{\bm x_i^t, \bm \tau_i^t\}_{i=1}^{n^t}$ of $n^t$ unlabeled SITS. 
\blue{We assume both the source and target domains consist of SITS acquired over a single year (January 1 to December 31) and in geographically different locations.}
The two domains can be associated with different data distributions, as changes in local conditions, \eg soil, weather, climate, or farmer practices, cause domain discrepancies~\cite{tuia2016domain}.
\green{In this work, we focus on the domain discrepancies caused by temporal shifts (Section~\ref{sec:introduction}).
Although not explicitly addressed in this work, there are other sources of discrepancies that might occur. For example, the local topography or soil conditions could impact not only the temporal development of crop growth, but also the spectral values, which could change the spectral signature of the same crop type in different regions.}
%which could cause changes in the spectral signatures of the same crop types.}

% 

%but also shift the spectral values, which could cause the spectral signatures of the same crop types to differ.}

Because of these data discrepancies, models which are trained  with the labeled source domain can fail when applied to the unlabeled target domain~\cite{lucas2021bayesian}, thus hindering the large-scale application of crop classifiers. To this end, our goal is to adapt a model trained on $\D^s$ to make accurate predictions on $\D^t$.
\green{We do so by explicitly estimating the temporal shift between the two regions to generate accurate pseudo-labels for $\D^t$. Then, we re-train the model with target data using the pseudo-labels, thereby adapting the model to the spectral and temporal properties of the target region.}     
We note that the classes in the source may not be exactly the same as the classes in the target.
This complicates UDA, which typically assumes a closed-set setting~\cite{panareda2017open}, where the set of classes in the source and target domains are equal. For simplicity, we focus on a closed-set setting by adapting a classifier trained for the main $K-1$ crop types in the source region, plus an ``unknown" class containing all remaining source data. This ensures that all target examples can be classified to either one of the $K-1$ crop classes or ``unknown".

%\green{Although not explicitly addressed in this work, there are other sources of domain shift that might occur. For example, the local topography or soil conditions could impact the spectral signatures of the same crop in two different geographical areas while the timing remains the same.}
%
%This causes spectral and temporal shifts, as the local topography or soil conditions could impact the spectral signatures of crops, while the local climate impacts the temporal shift of 
%
%in two different geographical areas while the timing remains the same. The local climate could change the temporal     
%
%spectral and temporal shifts.
%Spectral shifts may be caused by changes in topography or the soil, which impacts the spectral signatures of the same crop, while differences in climate could change           

%even if the peaks of NDVI occur at the same time, the maximum NDVI value will be different.}
% peak of growth / the maximum NDVI value / the amplitude will differ. 

\subsection{Approach Overview}
Here we give an overview of how TimeMatch addresses the cross-region UDA problem before describing the full details.
A visual presentation of TimeMatch is given in Figure~\ref{fig:methods_figure}.
TimeMatch consists of two components (i) temporal shift estimation and (ii) TimeMatch learning.

We aim to estimate the temporal shift between the source and target regions to reduce their domain discrepancy (see Figure~\ref{fig:ndvi}). We represent the temporal shift by a scalar $\delta^{t \to s} \in \mathbb{Z}$ (as the number of days), here in the direction from target to source. 
Note that the shift in the opposite direction is obtained by $\delta^{s \to t} = -\delta^{t \to s}$, so we only have to estimate one shift.
To shift the target domain by $\delta^{t \to s}$, we write $\bm \tau^t + \delta^{t \to s}$,
meaning $\delta^{t \to s}$ is added element-wise to each target day-of-year.
With our proposed method for temporal shift estimation (Section~\ref{sec:temporalshiftestimation}), we obtain estimates for 
$\delta^{t \to s}$ and $\delta^{s \to t}$.

In TimeMatch learning (Section~\ref{sec:timematchlearning}), we use $\delta^{s \to t}$ to construct a target-shifted source domain $\D^{s \to t} = \{(\bm x_i^s, \bm \tau_i^s + \delta^{s \to t}), y_i^s\}_{i=1}^{n^s}$, which has reduced domain discrepancy to the unlabeled target domain $\D^t$ due to the temporal alignment. We therefore use self-training to learn from the labeled $\D^{s \to t}$ and unlabeled $\D^t$. To do so, TimeMatch learning unifies temporal shift estimation with the loss function of FixMatch~\cite{sohn2020fixmatch} and the exponential moving average (EMA) training of Mean Teacher~\cite{tarvainen2017mean}, as we explain next.

We first obtain source-trained parameters by training a crop classifier with $\D^s$.
We then duplicate the trained classifier into two models: the \emph{teacher} and the \emph{student}.
Our TimeMatch learning algorithm aims to adapt both the teacher and the student to the new target region \blue{with self-training}.
The teacher generates pseudo-labels for the target domain to train the student, and the knowledge learned by the student is then updated back to the teacher, thus the pseudo-labels used to train the student itself are improved. 
We generate pseudo-labels by using $\delta^{t \to s}$ to create an adapted target domain $\D^{t \to s} = \{\bm x_i^t, \bm \tau_i^t + \delta^{t \to s}\}_{i=1}^{n^t}$. As $\D^{t \to s}$ is temporally aligned with $\D^s$, the source-initialized teacher generates more accurate pseudo-labels for $\D^{t \to s}$ than $\D^t$. 
The student is then trained with labeled $\D^{s \to t}$ and pseudo-labeled $\D^t$ via the FixMatch loss~\cite{sohn2020fixmatch}, 
thereby leveraging both the available source labels and the target pseudo-labels to adapt the student to the target domain.

After updating the student, the teacher is updated via an EMA of the student parameters.
As the two models adjust to the temporal shift of the target domain, 
the best shift $\delta^{t \to s}$ for pseudo-labeling with the teacher gradually moves to zero during TimeMatch learning.
To adjust to the changing shift and ensure the pseudo-labels are consistently accurate, it is necessary to re-estimate the temporal shift of the teacher as it learns.
However, repeating temporal shift estimation is computationally expensive, and drastically increases training time if done each training iteration. Therefore, in Section~\ref{sec:ema}, we discuss how EMA training alleviates this issue by enabling the re-estimation to be done only once per epoch.

Next, we first describe our method for estimating the temporal shift before describing the loss function and learning algorithm of TimeMatch learning.

\subsection{Temporal Shift Estimation}
\label{sec:temporalshiftestimation}
Estimating the temporal shift directly from the data is difficult, as labels are not available in the target domain. Without labels, we cannot separate the target data into each crop type, which prevents the computation of \eg vegetation indices to compare the source and target phenology of each crop type directly.

Instead, we propose to estimate the temporal shift by calculating statistics on the predictions of a source-trained model when input temporally shifted target data.
By doing so, we estimate the shift that aligns the target data with the source crop phenology learned by a model, 
leveraging the classification ability of the trained model to estimate the shift from unlabeled data.
Another benefit of this approach is that it enables re-estimation of the best temporal shift for pseudo-labeling as the learned phenology of the model changes from source to target in TimeMatch learning.

One possible value to measure is the confidence of the model predictions. Intuitively, when a source-trained model is applied to correctly shifted target data, it should output more confident predictions than for incorrectly shifted target data. As correctly classified examples tend to have more confident predictions than wrongly classified or out-of-distribution examples~\cite{hendrycks2016baseline}, we argue that a confident temporal shift indicates a better alignment of the target domain with the source which results in accurate pseudo-labels and reduced domain discrepancy.

\blue{The confidence of a model for a particular shift $\delta^{t \rightarrow s}$ can be measured by the expected entropy:}
\begin{equation}
    \label{eq:entropy}
    \E_{(\bm x^t, \bm \tau^t) \sim \D^t} \left[H\mleft(p_\theta\mleft(y | (\bm x^t, \bm \tau^t + \delta^{t \rightarrow s})\mright)\mright)\right],
\end{equation}
where $H$ denotes the entropy, here computed over the predictions of the model $\theta$ when input temporally shifted target data sampled from $\D^t$. 

\blue{To estimate a temporal shift with entropy, Equation~\ref{eq:entropy} should be computed for each possible shift \mbox{$\delta^{t \to s} \in \{-\Delta, -\Delta+1, \dots, \Delta\}$}, and the estimated shift is then the one with lowest entropy. Here, $\Delta$ defines the maximum possible shift (in days) to estimate between the source and target regions. }

\blue{However, due to the class imbalance of SITS, relying on expected entropy alone could result in choosing a shift where the model outputs confident predictions for only the most frequent classes while ignoring the less frequent classes. This would hinder the adaptation of the model for the less frequent target classes.
 To address this problem, the diversity of the predicted marginal distribution should also be considered in the estimation. The marginal is given by:}
\begin{equation}
    p_\theta(y) = \E_{(\bm x^t, \bm \tau^t) \sim \D^t} \left[p_\theta\mleft(y | \left(\bm x^t, \bm \tau^t + \delta^{t \rightarrow s}\right)\mright)\right],
\end{equation}
that is, the expected predictions of the model (parameterized by $\theta$) when input shifted target data. 

Ideally, the marginal distribution should match the class distribution of the target domain,
as this indicates a shift where the model predicts a diverse set of classes according to their actual frequency.
But since target labels are unavailable, so is the target class distribution.
Instead, inspired by metrics for evaluating image generative models, we consider two options to address this: 
the Inception score~\cite{salimans2016improved} (IS), and the activation maximization score~\cite{zhou2017activation} (AM).
Both metrics consider the entropy and marginal of a pre-trained model, but IS scores the marginal distribution by its similarity to a uniform distribution, whereas AM uses the actual class distribution.

As these metrics were originally proposed to evaluate the quality of generated images, we describe next how we repurpose them for temporal shift estimation. Finally, we describe an algorithm where IS is used to initialize the temporal shift for estimating the target class distribution with pseudo-labels and enable a better temporal shift estimate with AM.

\subsubsection{Inception Score}
IS is computed for a temporal shift $\delta$ by:
\begin{align}
    \label{eq:inception_score}
    &\mathrm{IS}(\delta^{t \to s}, \theta) \nonumber \\
    &= \E_{(\bm x^t, \bm \tau^t)}
    \left[\kld*{
        p_\theta\mleft(y | (\bm x^t, \bm \tau^t + \delta^{t \to s})\mright)
    }{
        p_{\theta}(y)
    }\right] \\
    &= H(p_\theta(y)) 
    - \E_{(\bm x^t, \bm \tau^t)}\left[H\mleft(p_\theta\mleft(y | (\bm x^t, \bm \tau^t + \delta^{t \to s})\mright)\mright)\right]
\end{align}
where $\kld{\cdot}{\cdot}$ is the KL-divergence between two distributions, here the conditional distribution $p_\theta\big(y | (\bm x^t, \bm \tau^t + \delta)\big)$ and marginal distribution $p_{\theta}(y)$ predicted with model parameters $\theta$. Higher values of IS indicate a better $\delta$, as when the conditional and marginal distributions are different, this corresponds to a temporal shift where the former has low entropy (\ie, the model is confident), and the latter has high entropy (\ie, the model predicts a diverse set of classes). Hence, the temporal shift $\delta^{t \to s}$ is estimated by:
\begin{equation}
    \delta_{IS}^{t \rightarrow s}(\theta^s) = \argmax_{\delta^{t \to s} \in \{-\Delta, \dots, \Delta\}} \mathrm{IS}(\delta^{t \to s}, \theta^s),
    \label{eq:inception_shift}
\end{equation}
where the estimated temporal shift maximizes IS for a source-trained model parameterized by $\theta^s$ when applied to target data.

\subsubsection{AM Score}
A shortcoming of IS is that the highest score is achieved when $p_\theta(y)$ is uniform~\cite{barratt2018note}, which corresponds to an even distribution of classes in the target domain. For SITS, where the class distribution is often highly imbalanced, this may cause IS to estimate a suboptimal shift. AM~\cite{zhou2017activation} addresses this issue by taking the actual target class distribution $C^t$ into account:
\begin{equation}
\begin{split}  
     \mathrm{AM}(\delta^{t \to s}, \theta, C^t)
    & = \mathbb{E}_{(\bm x^t, \bm \tau^t)} \left[H\mleft(p_\theta\mleft(y | (\bm x^t, \bm \tau^t + \delta^{t \to s})\mright)\mright)\right] \\
    & + \kld{C^{t}}{p_\theta(y)}.
    \label{eq:am_score}
\end{split}
\end{equation}
AM consists of two terms: the first term is an entropy term on the conditional distribution, 
and the second is the KL-divergence between the underlying class distribution $C^t$ and the marginal distribution. Lower values of AM indicate a better $\delta$, as the model is confident in its predictions, and the actual class distribution of the data matches the predicted distribution of classes. The temporal shift $\delta^{t \to s}$ is estimated by:
\begin{equation}
    \label{eq:am_shift}
    \delta_{AM}^{t \rightarrow s}(\theta^s, C^t) = \argmin_{\delta^{t \to s} \in \{-\Delta, \dots, \Delta\}} \mathrm{AM}(\delta^{t \to s}, \theta^s, C^t).
\end{equation}
where the estimated temporal shift minimizes AM. 

\subsubsection{Algorithm for Estimating Temporal Shift}
\blue{
While AM is more accurate at estimating the temporal shift, it requires knowledge of the target class distribution $C^{t}$, which is not available.
To address this, we propose to approximate the target class distribution for AM by pseudo-labels obtained with IS. We show our approach in Algorithm~\ref{alg:estimateshift}.
First, we use IS (Equation~\ref{eq:inception_shift}) to estimate an initial shift $\delta^{t \rightarrow s}$ (line 3). 
This initial estimate allows us to shift the target domain so that more accurate pseudo-labels can be generated with a source-trained model. We then use the pseudo-labels to estimate the target class distribution $\hat{C}^t$ (lines 4-5). 
Finally, we re-estimate the temporal shift more accurately with AM and $\hat{C}^t$ (line 6).
}

\begin{algorithm}[t]

\textbf{Input:} Source-trained parameters $\theta^s$, target domain $\D^t$, target class distribution estimate $\hat{C}^t$

\If{$\hat{C}^t = \bm{0}$} {
    Estimate temporal shift $\delta^{t \to s} \gets \delta_{IS}^{t \to s}(\theta^s)$ 
    (Eq.~\ref{eq:inception_shift})

    Compute pseudo labels for each $\left(\bm x_i^t, \bm \tau_i^t\right) \in \D^t$:
    $\hat{y}_i^t \gets \argmax_y\left(p_{\theta^s}(y | \bm x_i^t, \bm \tau_i^t + \delta^{t \rightarrow s})\right)$

    Estimate class distribution $\hat{C}^t_y \gets \frac{1}{n^t} \sum_{i=1}^{n^t}\bm{1}_{\hat{y}_i^t = y}$ for $y \in \{1,\dots,K\}$

}

Estimate temporal shift $\delta^{t \to s} \gets \delta_{AM}^{t \to s}(\theta^s, \hat{C}^t)$  (Eq.~\ref{eq:am_shift})

{\bfseries Output:} Temporal shift $\delta^{t \to s}$
\caption{\textsc{EstimateTemporalShift}}
\label{alg:estimateshift}
\end{algorithm}

\subsection{TimeMatch Learning}
\label{sec:timematchlearning}
\begin{algorithm*}[ht]

\textbf{Input:}
Labeled source domain $\mathcal{D}^s$, unlabeled target domain $\mathcal{D}^t$,
source-trained parameters $\theta^s$, total epochs $n$ and iterations $m$, pseudo label threshold $\epsilon$, trade-off value $\lambda$, EMA decay rate $\alpha$, learning rate $\eta$

Initialize student parameters $\theta \gets \theta^s$ and teacher parameters $\theta' \gets \theta^s$  \label{alg:timematch:init_model}

Initialize estimated target class distribution $\hat{C}^t = \bm{0}$

\For{epoch = 1 \textbf{to} $n$} {
    Estimate temporal shift with teacher: $\delta^{t \rightarrow s} \gets$ \textsc{EstimateTemporalShift}($\theta'$, $\D^t$, $\hat{C}^t$)  \label{alg:timematch:estimateshift}

    \If {epoch = 1} {
        Initialize $\delta^{s \rightarrow t} \gets -\delta^{t \rightarrow s}$  \label{alg:timematch:sourceshift}
    }

    \For{iteration = 1 \textbf{to} $m$} {

        Sample mini-batches of size $B$ from source $\mathcal{S} = \{(\bm x_i^s, \bm \tau_i^s, y_i^s)\}_{i=1}^B$ and target $\mathcal{T} = \{(\bm x_i^t, \bm \tau_i^t)\}_{i=1}^B$

        With $\mathcal{S}$ shifted by $\delta^{s \to t}$, compute source loss $\mathcal{L}^{s \rightarrow t}$ (Eq.~\ref{eq:loss_src_to_trg}) \label{alg:timematch:src_loss}

        For each example in $\mathcal{T}$ shifted by $\delta^{t \to s}$, generate teacher prediction $\bm q_i^t$ and pseudo labels $\hat{y}_i^t$  (Eq.~\ref{eq:pseudo_preds} and~\ref{eq:pseudo_label}) \label{alg:timematch:pseudo_label}

        With $\mathcal{T}$ and confident pseudo labels $\hat{y}_i^t$ with $\max(\bm q_i^t) > \epsilon$, compute target loss $\mathcal{L}^{t}$  (Eq.~\ref{eq:loss_trg}) \label{alg:timematch:trg_loss}

    Update student by gradient: $\theta \gets \theta - \gamma\nabla_\theta (\mathcal{L}^{s \to t} + \lambda\mathcal{L}^{t})$ \label{alg:timematch:update_student}

    Update teacher by EMA: $\theta' \gets (1-\alpha)\theta + \alpha \theta'$  \label{alg:timematch:update_teacher}

    }

    Re-estimate class distribution: $\hat{C}^t_y \gets \frac{1}{mB} \sum_i \bm{1}_{\hat{y}_i^t = y}$ for $y \in \{1,\dots,K\}$ (using all pseudo labels from epoch)
}
{\bfseries Output:} Student parameters $\theta$
\caption{\textsc{TimeMatch}}
\label{alg:timematch}
\end{algorithm*}

With our method for estimating the temporal shift, we can reduce the domain discrepancy between the source and target domains.
The TimeMatch learning algorithm uses the temporal shift to train the student model for the target domain from teacher-generated pseudo-labels via the FixMatch loss~\cite{sohn2020fixmatch} and EMA training~\cite{tarvainen2017mean}. We present the complete TimeMatch algorithm in Algorithm~\ref{alg:timematch}, and describe the details of each step in the following.

\subsubsection{Pre-training on the Source Domain} 
As we rely on the teacher to generate pseudo-labels to train the student, it is important to obtain a good initialization for both models. Additionally, temporal shift estimation requires a source-trained model.
Thus, we first use the labeled source domain to obtain source-trained model parameters $\theta^s$.
Given a batch of labeled source data from $\D^s$, we optimize the following loss function:
\begin{equation}
\mathcal{L}^s = \frac{1}{B} \sum_{i=1}^{B} L\mleft(p_{\theta^s}\big(y | (\bm x_i^s, \bm \tau_i^s )\big), y_i^s\mright),
\end{equation}
where $L(\cdot, \cdot)$ is a classification loss (\eg cross-entropy or focal loss~\cite{lin2017focal}) and $B$ the batch size. After pre-training, we initialize the parameters of the student $\theta$ and teacher $\theta'$ from $\theta^s$ (line~\ref{alg:timematch:init_model}).

\subsubsection{TimeMatch Loss}
The TimeMatch loss consists of two terms: a supervised loss $\mathcal{L}^{s \rightarrow t}$ applied to the adapted source domain $\D^{s \to t}$ and an unsupervised loss $\mathcal{L}^{t}$ applied to the unlabeled target domain $\D^t$. Our loss is based on the FixMatch loss~\cite{sohn2020fixmatch}. 
To regularize the model to predict consistent pseudo-labels on randomly augmented versions of the same inputs, FixMatch applies two types of augmentation functions: \textit{weakly}-augmented $a(\cdot)$ and \textit{strongly}-augmented $A(\cdot)$, corresponding to simple and extensive augmentations of the input.
We describe the form of augmentations we use for $a(\cdot)$ and $A(\cdot)$ in Section~\ref{sec:impl}.

Let $\delta^{s \to t}$ and $\delta^{t \to s}$ be temporal shifts estimated given by Algorithm~\ref{alg:estimateshift} using the teacher (line~\ref{alg:timematch:estimateshift}-\ref{alg:timematch:sourceshift}). To compute the supervised loss on the source domain, we use $\delta^{s \rightarrow t}$ to align the source domain with the target domain and optimize:
\begin{equation}
    \label{eq:loss_src_to_trg}
    \mathcal{L}^{s \rightarrow t} = \frac{1}{B} \sum_{i=1}^{B} L\mleft(p_\theta\big(y | A(\bm x_i^s, \bm\tau_i^s + \delta^{s \rightarrow t})\big), y_i^s\mright),
\end{equation}
using source labels $y_i^s$ to update the student $\theta$ on strongly augmented source data shifted by $\delta^{s \to t}$. This loss makes it possible for the student to learn the target phenology from shifted source data (line~\ref{alg:timematch:src_loss}).

To generate pseudo-labels for the target domain, we obtain the predicted class distribution from the teacher when input source-shifted target data:
\begin{equation}
    \bm q_i^t = p_{\theta'}\big(y | a\mleft(\bm x_i^t, \bm \tau_i^t + \delta^{t \rightarrow s}\mright)\big),
\label{eq:pseudo_preds}
\end{equation}
where the teacher $\theta'$ is input a weakly-augmented target sample, shifted by $\delta^{t \rightarrow s}$. Then, we use
\begin{equation}
    \hat{y}_i^t = \argmax (\bm q_i^t)
\label{eq:pseudo_label}
\end{equation}
as pseudo-label (line~\ref{alg:timematch:pseudo_label}). The student $\theta$ is then updated on 
strongly-augmented target data for confident pseudo-labels (line~\ref{alg:timematch:src_loss}):
\begin{equation}
    \label{eq:loss_trg}
    \mathcal{L}^{t} = \frac{1}{B} \sum_{i=1}^{B} 
    \bm{1}_{\max(\bm q_i^t) > \epsilon}
    L\mleft(p_\theta\big(y | A\mleft(\bm x_i^t, \bm \tau_i^t\mright)\big), \hat{y}_i^t\mright),
\end{equation}
where $\bm{1}$ is the indicator function, and $\epsilon$ is the confidence threshold for using a pseudo-label. With this loss, the student is trained with target data using pseudo-labels.
The total loss minimized by the student in TimeMatch is:
\begin{equation}
\mathcal{L}^{all} = \mathcal{L}^{s \rightarrow t} + \lambda \mathcal{L}^t,
\label{eq:timematch_loss}
\end{equation}
where $\lambda$ is a scalar hyperparameter to control the trade-off between the supervised and the unsupervised loss (line~\ref{alg:timematch:update_student}).

\subsubsection{EMA training and re-estimating temporal shift}
\label{sec:ema}
By optimizing $\mathcal{L}^{all}$, the student and teacher are trained only for the target phenology, 
as $\mathcal{L}^{s\rightarrow t}$ shifts the time of the source to the target, while $\mathcal{L}^t$ keeps the target in its original time. This loss enables a source-trained model to adapt to the crop phenology of the target domain.

However, by doing so, the source domain is gradually ``forgotten", and as a result, it becomes unnecessary to apply the temporal shift $\delta^{t \to s}$ for pseudo-labeling the target domain with the teacher. This causes $\delta^{t \to s}$ to gradually move to zero during TimeMatch learning. 
Thus, if $\delta^{t\to s}$ is fixed to the same shift, the target samples will be wrongly shifted, which results in incorrect pseudo-labels. To address this, we re-estimate the temporal shift for the teacher during TimeMatch learning.
As Algorithm~\ref{alg:estimateshift} chooses the shift based on the confidence and diversity of model predictions, re-estimating the temporal shift with the teacher ensures the generated pseudo-labels remain accurate during training.

However, if the teacher is a direct copy of the student, the model will rapidly adapt to the target domain, which requires the temporal shift to be re-estimated every few iterations. But doing so drastically increases training time, as Equation~\ref{eq:am_shift} requires forwarding a large sample of target data for each possible temporal shift.
We address this by introducing EMA training, where the teacher is slowly updated via an EMA of the student parameters (line~\ref{alg:timematch:update_teacher}):
\begin{equation}
    \theta' \leftarrow (1-\alpha)\theta + \alpha \theta',
\end{equation}
where $\alpha$ is a decay rate. By choosing $\alpha$ close to 1, 
we reduce the rate at which the teacher adapts to the target domain, enabling the 
re-estimation of $\delta^{t \to s}$ to be done only once each epoch (line~\ref{alg:timematch:estimateshift}).
Moreover, by averaging model weights via EMA, we also obtain less noisy pseudo-labels~\cite{tarvainen2017mean}.

\begin{figure*}[ht]
\centering
\includegraphics[width=0.8\linewidth]{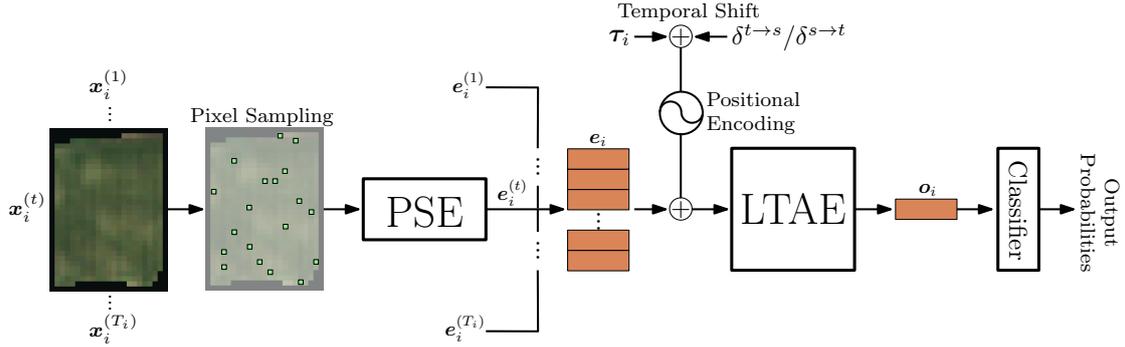}
\caption{\blue{Overview of the PSE+LTAE model~\cite{garnot2020satellite, garnot2020lightweight}. Given SITS of an agricultural parcel, the PSE module process each time step independently by embedding a random sample of pixels. The results are then concatenated into a sequence of embeddings $\bm{e}_i$.
The observation dates $\bm{\tau}_i$, which we add temporal shifts to, are input to the model by adding their positional encoding to $\bm{e}_i$. The result is temporally processed by LTAE to a single embedding $\bm{o}_i$ which is then passed to the classifier.}} 
\label{fig:psetae}
\end{figure*}
By re-estimating the temporal shift, the teacher and the shift can both evolve jointly during training, resulting in better pseudo-labels for improved cross-region adaptation. 
Note that $\delta^{s \to t}$ is not re-estimated (line~\ref{alg:timematch:sourceshift}). 
The first shift estimate represents the shift of the data, whereas the re-estimated shift represents the shift of the teacher.
By fixing $\delta^{s \to t}$ to the initial estimate, the source domain is kept aligned with the target domains during training, which enables semi-supervised learning.

\section{Dataset and Materials}
\label{sec:data_and_material}
This section presents the TimeMatch dataset~\cite{nyborg2021timematchdataset} and the materials for our experiments. We first introduce the crop classification model we use, followed by a description of the dataset and its pre-processing.
Then, we describe the competitors and our implementation. Our source code is publicly available, and contains the implementation of TimeMatch and the competitors, a link to download our dataset, and the full experimental results: \github.

\subsection{Network Architecture}
As model, we use the existing object-based crop classifier PSE+LTAE introduced by Sainte Fare Garnot~\etal~\cite{garnot2020satellite, garnot2020lightweight}.
The network consists of two modules: the pixel-set encoder (PSE) and the lightweight temporal attention encoder (LTAE). \blue{See Figure~\ref{fig:psetae} for an overview.}

The PSE module handles the spatial and spectral context of SITS. \blue{Given SITS of an agricultural parcel, PSE samples a random pixel-set of size $S$ among the $N_i$ available pixels within the parcel.} The PSE is efficient compared to \eg convolutions, which are time and memory-consuming when applied to irregularly sized parcels. As spatial information is lost by doing so, the PSE supports an optional extra input with various geometrical properties of the given parcel, such as its area. We do not input this extra feature to avoid biasing the model towards the shapes of parcels in the source region, which typically change depending on the local farmer practices. 
Thus, we only input the sequence $\bm x_i \in \R^{T_i \times N_i \times C}$, which is then embedded by the PSE for each time step independently. 

The LTAE module~\cite{garnot2020lightweight} handles the temporal context by applying self-attention~\cite{vaswani2017attention} with modifications to output a single embedding. It improves the accuracy and computational efficiency compared to the original TAE~\cite{garnot2020satellite} by a channel grouping strategy and a learnable master query.
The additional input $\bm \tau_i$ is input to LTAE by encoding the days with \blue{sinusoidal positional encoding~\cite{vaswani2017attention}} and adding the result to the output of PSE. 
As the positional encoding does not support negative inputs, we input negative temporal shifts by offsetting each $\bm \tau_i$ by the maximum temporal shift $\Delta$. Given the sequence of PSE-embeddings and the encoded $\bm \tau_i$, LTAE outputs a single embedding $\bm o_i$, which is then classified by a multi-layer perceptron to produce class probabilities $p(y | (\bm x_i, \bm \tau_i)) \in \R^K$.

\subsection{The TimeMatch Dataset}
\begin{figure}[ht]
\centering
\frame{\includegraphics[width=0.9\linewidth]{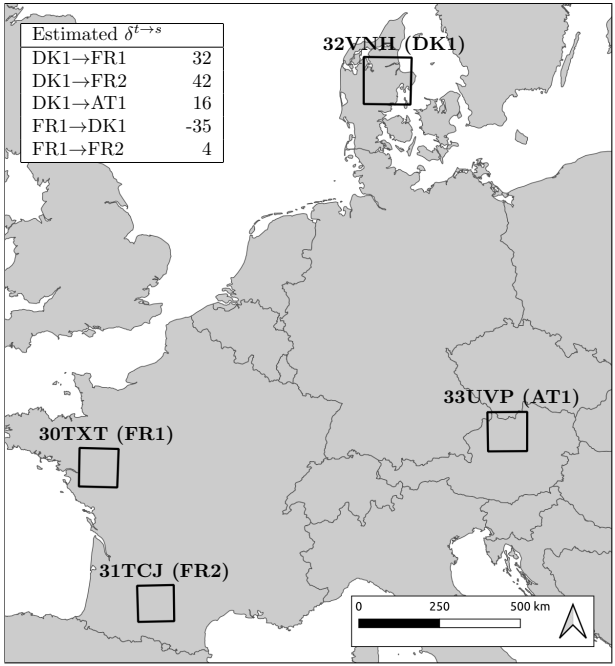}}
\caption{Locations of the four European Sentinel-2 tiles in the TimeMatch dataset.
In the upper left corner, we show the temporal shifts $\delta^{t \to s}$ estimated by Algorithm~\ref{alg:estimateshift} with a source-trained model.}
\label{fig:tile_map}
\end{figure}

\begin{figure}[ht]
\centering
\includegraphics[width=1.0\linewidth,trim={0.5cm 1.5cm 0.5cm 0cm}]{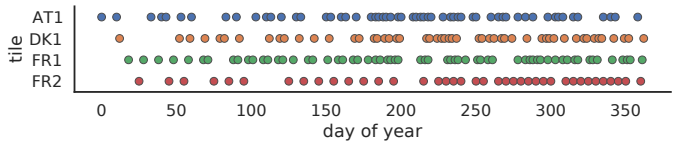}
\caption{Acquisition dates for each Sentinel-2 tile in our dataset. The inputs are irregularly sampled with variable temporal length.}
\label{fig:acquisition_dates}
\end{figure}

The TimeMatch dataset~\cite{nyborg2021timematchdataset} contains SITS from Sentinel-2 Level-1C products in top-of-atmosphere reflectance.
Four Sentinel-2 tiles are chosen in various climates: 33UVP (Austria), 32VNH (Denmark), 30TXT (mid-west France), and 31TCJ (southern France), \blue{abbreviated as AT1, DK1, FR1, and FR2, respectively.}
A map of the tiles is shown in Figure~\ref{fig:tile_map}.
We use all available observations with cloud coverage $\leq80\%$ and coverage $\geq 50\%$ between January 2017 and December 2017. Figure~\ref{fig:acquisition_dates} shows the resulting acquisition dates for the four tiles.
We leave out the atmospheric bands (1, 9, and 10), keeping $C=10$ spectral bands. The 20m bands are bilinearly interpolated to 10m. 

\begin{figure}[ht]
\centering
\includegraphics[width=0.9\linewidth,trim={1cm 1cm 1cm 1cm}]{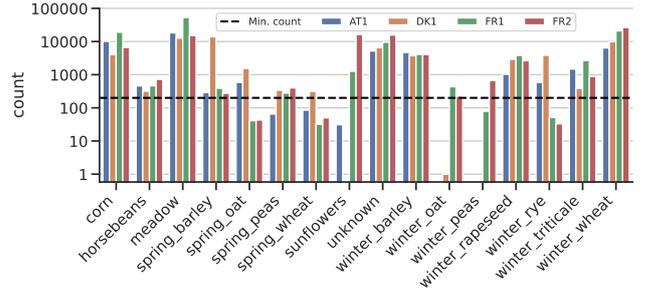}
\caption{Class frequencies (log scale) for each Sentinel-2 tile in the TimeMatch dataset. The dashed line indicates the threshold for the source region when selecting a class as part of the $K$ classes.}
\label{fig:class_distribution}
\end{figure}

For ground truth data, we retrieve geo-referenced parcel shapes and their crop type labels from the openly available Land Parcel Identification System (LPIS) records in Denmark\footnote{https://kortdata.fvm.dk/download (``Marker")}, France\footnote{http://professionnels.ign.fr/rpg (``RPG")}, and Austria\footnote{https://www.data.gv.at (``INVEKOS Schläge")}.
We select 15 major crop classes in Europe and label any remaining parcels as unknown.
Figure~\ref{fig:class_distribution} shows the selected classes and their frequency in each tile.

We pre-process the parcels by applying 20m erosion and removing all parcels with an area of less than 1 hectare. This reduces label noise by removing pixels near the border of parcels, which are often less representative of the given crop class compared to the pixels in the middle, and also by removing small or thin polygons, which are typically miscellaneous classes such as field borders.
The SITS are pre-processed for object-based classification by cropping the pixels within each parcel to input sequences $\bm x_i \in \R^{T_i \times N_i \times 10}$. 
Each input is then randomly assigned to the train/validation/test sets of each Sentinel-2 tile by a 70\%/10\%/20\% ratio.
Note that this process assumes knowledge of parcel shapes in the target region. If this is not available, TimeMatch may instead be applied for pixel-based classification by inputting single pixels ($S=1$) to PSE+LTAE. 
We choose five different cross-region tasks (written as ``source$\to$target"): DK1$\to$FR1, DK1$\to$FR2, DK1$\to$AT1, FR1$\to$DK1, and FR1$\to$FR2. 
When a Sentinel-2 tile is chosen as source, all labels of the train and validation sets are available for training. When a tile is the target region, no labels are available, except for the final evaluation on the test set.

In contrast, many existing UDA methods assume that a labeled validation set is available for the target domain, and use it during training \eg to select the best model~\cite{ganin2016domain, long2017conditional, tzeng2014deep, chen2020adversarial}. 
However, this assumption is unrealistic, as if labels were available in real-world scenarios, they would be better used for training the model. Instead, we report all cross-region UDA test results with the model output at the end of training.
Still, hyperparameters must be chosen with a labeled validation set. Thus, we tune hyperparameters with the target validation set for only one task, DK1$\to$FR1, and apply the found hyperparameters to all remaining tasks (as done in~\cite{fatras2021jumbot}).

The class distributions between regions differ significantly, and there may not be enough examples of a crop type in the source region for a model to learn their classification.
Thus, when pre-training models on source data, we only use a subset of the available crop types with at least 200 examples in the source region (as indicated by the dashed line in Figure~\ref{fig:class_distribution}).
The remaining classes are set as ``unknown". When evaluating on the target data, we report results on the same selection of source classes no matter their frequency in the target.

\subsection{Comparisons}
\label{sec:comparisons}

\paragraph{Baselines} We consider the following baseline methods:
\begin{itemize}
    \item \emph{Source-Trained} is PSE+LTAE trained on the source domain and applied to the target domain without domain adaptation. This result represents the lower bound cross-region performance of the model.
    \item \emph{Target-Trained} is PSE+LTAE trained with labeled target data using the same classes as the source-trained. We note that by training with the source classes, which is required for comparison, infrequent classes may not be learned properly which increases the variance of the results. This result can be seen as the upper bound cross-region performance if labels were available in the target region.
\end{itemize}

\paragraph{Competing UDA Methods}
We compare TimeMatch to five existing UDA methods. We reproduce these methods for SITS by replacing the original feature extractor with PSE+LTAE. 
For domain-invariant methods, we align the LTAE feature vector input to the final classifier \blue{(\ie, $\bm{o}_i$ in Figure~\ref{fig:psetae})}, similar to the original approach in these methods.

We compare to the following methods:
\begin{itemize}
    \item \emph{FixMatch}~\cite{sohn2020fixmatch} is TimeMatch without the temporal shift estimation.  
    As this method is semi-supervised learning, it shows whether UDA or SSL is more beneficial for cross-region adaptation.
    \item \emph{MMD}~\cite{tzeng2014deep} learns domain-invariant features by minimizing the maximum mean discrepancy metric.
    \item \emph{DANN}~\cite{ganin2015unsupervised} uses a domain classifier to learn domain-invariant features with adversarial training. 
    \item \emph{CDAN+E}~\cite{long2017conditional} improves upon DANN by conditioning the domain classifier on the classification output and minimizing an entropy loss on target data.
    \item \blue{\emph{ALDA}~\cite{chen2020adversarial} is a self-training method where pseudo-labels are refined by a noise-correcting domain discriminator. This method is in essence the most similar to TimeMatch.}
    \item \emph{JUMBOT}~\cite{fatras2021jumbot} learns domain-invariant features by a discrepancy measure based on optimal transport. 
\end{itemize}
We note that time-series domain adaptation methods R-DANN, VRADA~\cite{purushotham2016vrada} and CoDATS~\cite{wilson2020codats} also employ DANN to align the features extracted by temporal network architectures. Thus, the only difference between VRADA, CoDATS, and the DANN approach mentioned here is the backbone architecture, which in our case is the temporal model PSE+LTAE.

PAN~\cite{wang2021phenology}, a UDA method for SITS,
learns domain-invariant features by minimizing the MMD loss for a temporal crop classification network. Unfortunately, we were unable to gain access to the source code of PAN for comparison. As an alternative, we include the MMD comparison, which is similar to PAN, except the crop classifier is changed to PSE+LTAE. 

\paragraph{ShiftAug} \blue{To verify the benefits of estimating the temporal shift compared to training models that are invariant to temporal shifts, we implement a simple data augmentation technique to train shift-invariant models that we name \textit{ShiftAug}. During training, ShiftAug uniformly samples $\delta \sim \mathcal{U} (-\Delta, \Delta)$ for each training example and shifts the example by $(\bm{x}_i, \bm{\tau}_i + \delta)$. By extending the training data to contain all valid temporal shifts with uniform probability, ShiftAug enables training models with invariance towards shifts.
Note that ShiftAug is incompatible with the temporal shift estimation presented in Algorithm~\ref{alg:estimateshift}, which requires a shift-variant model.}

\blue{We implement all competing methods with and without ShiftAug. This reveals the degree at which existing methods can implicitly learn shift-invariance.}

\subsection{Implementation Details}
\label{sec:impl}
All experiments are implemented in PyTorch~\cite{paszke2019pytorch} and trains on a single NVIDIA 1080 Ti GPU. Our implementation is based on the source code of PSE+LTAE~\cite{garnot2020lightweight}.

\paragraph{Source-training}
To initialize models on the labeled source domain, we follow the original training approach of PSE+LTAE~\cite{garnot2020satellite}.
We train for 100 epochs with the Adam~\cite{kingma2014adam} optimizer with an initial learning rate of $0.001$ and we decay the learning rate using a cosine annealing schedule~\cite{loshchilov2016sgdr}. We use weight decay of $0.0001$, batch size $128$, and focal loss $\gamma=1$. Inputs are normalized to $[0, 1]$ by dividing by the max 16-bit pixel value $2^{16}-1$. The best source-trained model is selected using the source validation set. We augment the inputs by randomly sub-sampling $30$ time steps. The pixel-set size of PSE is set to $S=64$ during training. 
The same setup is used for the target-trained model. 
For the final evaluation, we do not sample time steps or pixels, and instead input all available time steps ($T=T_i$) and pixels ($S=S_i$) for each example to the model. This ensures deterministic test results, and we also observe slightly improved results by doing so.

\paragraph{ShiftAug}
\blue{When training with ShiftAug, all training data (both source and target) are randomly shifted during training as described in Section~\ref{sec:comparisons}. ShiftAug is disabled during evaluation.}

\begin{table*}[htbp]
\begin{center}
\begin{tabu}{lc cccccc}
\toprule 
Method & \blue{ShiftAug} & DK1$\to$FR1 &  DK1$\to$FR2 & DK1$\to$AT1 & FR1$\to$DK1 & FR1$\to$FR2 & Avg. \\ 
\midrule
Source-trained & \xmark & 28.3$\pm$1.9 & 29.0$\pm$5.2 & 43.4$\pm$4.0 & 24.9$\pm$2.0 & 70.3$\pm$1.9 & 39.2$\pm$3.0 \\
%\rowfont{\color{blue}} & \cmark & 40.9$\pm$0.8 & 37.4$\pm$2.3 & 48.9$\pm$2.8 & 47.3$\pm$1.9 & 70.5$\pm$1.1 & 49.0$\pm$1.8 \\ 
 & \cmark & 40.9$\pm$0.8 & 37.4$\pm$2.3 & 48.9$\pm$2.8 & 47.3$\pm$1.9 & 70.5$\pm$1.1 & 49.0$\pm$1.8 \\ 
 \midrule
FixMatch~\cite{sohn2020fixmatch} & \xmark & 24.2$\pm$4.0 & 28.2$\pm$6.9 & 37.4$\pm$5.6 & 26.2$\pm$1.8 & 70.4$\pm$0.9 & 37.3$\pm$3.8 \\ 
%\rowfont{\color{blue}} & \cmark & 48.2$\pm$1.3 & 44.2$\pm$3.2 & 57.4$\pm$2.2 & 51.3$\pm$1.6 & 67.7$\pm$0.2 & 53.7$\pm$1.7 \\
 & \cmark & 48.2$\pm$1.3 & 44.2$\pm$3.2 & 57.4$\pm$2.2 & 51.3$\pm$1.6 & 67.7$\pm$0.2 & 53.7$\pm$1.7 \\
\midrule
MMD~\cite{tzeng2014deep} & \xmark & 36.6$\pm$0.7 & 35.5$\pm$0.6 & 49.7$\pm$2.0 & 32.5$\pm$2.0 & 61.6$\pm$2.6 & 43.2$\pm$1.6 \\
%\rowfont{\color{blue}} & \cmark & 42.2$\pm$0.4 & 39.5$\pm$0.8 & 48.9$\pm$2.4 & 42.8$\pm$2.3 & 59.0$\pm$2.7 & 46.5$\pm$1.7 \\ \cmidrule(lr){2-2}
 & \cmark & 42.2$\pm$0.4 & 39.5$\pm$0.8 & 48.9$\pm$2.4 & 42.8$\pm$2.3 & 59.0$\pm$2.7 & 46.5$\pm$1.7 \\ \cmidrule(lr){2-2}

DANN~\cite{ganin2015unsupervised} & \xmark & 38.7$\pm$0.7 & 37.3$\pm$0.6 & 52.0$\pm$1.4 & 34.0$\pm$1.8 & 71.0$\pm$0.2 & 46.6$\pm$0.9 \\
%\rowfont{\color{blue}} & \cmark & 45.3$\pm$2.2 & 44.1$\pm$1.4 & 52.4$\pm$1.4 & 42.9$\pm$2.5 & 68.7$\pm$0.5 & 50.7$\pm$1.6  \\ \cmidrule(lr){2-2}
 & \cmark & 45.3$\pm$2.2 & 44.1$\pm$1.4 & 52.4$\pm$1.4 & 42.9$\pm$2.5 & 68.7$\pm$0.5 & 50.7$\pm$1.6  \\ \cmidrule(lr){2-2}

CDAN+E~\cite{long2017conditional} & \xmark & 39.3$\pm$0.6 & 37.9$\pm$0.3 & 51.5$\pm$2.9 & 36.5$\pm$1.3 & 71.7$\pm$0.6 & 47.4$\pm$1.1 \\
%\rowfont{\color{blue}} & \cmark & 46.5$\pm$2.3 & 45.2$\pm$1.3 & 55.0$\pm$1.3 & 46.9$\pm$0.5 & 70.7$\pm$1.3 & 52.9$\pm$1.3 \\ \cmidrule(lr){2-2}
 & \cmark & 46.5$\pm$2.3 & 45.2$\pm$1.3 & 55.0$\pm$1.3 & 46.9$\pm$0.5 & 70.7$\pm$1.3 & 52.9$\pm$1.3 \\ \cmidrule(lr){2-2}

%\rowfont{\color{blue}}ALDA~\cite{chen2020adversarial} & \xmark & 36.9$\pm$0.2 & 33.1$\pm$1.9 & 47.2$\pm$3.9 & 35.0$\pm$1.0 & 55.3$\pm$3.1 & 41.5$\pm$2.0 \\
ALDA~\cite{chen2020adversarial} & \xmark & 36.9$\pm$0.2 & 33.1$\pm$1.9 & 47.2$\pm$3.9 & 35.0$\pm$1.0 & 55.3$\pm$3.1 & 41.5$\pm$2.0 \\
%\rowfont{\color{blue}}& \cmark & 42.8$\pm$2.1 & 36.2$\pm$0.6 & 51.5$\pm$2.2 & 40.7$\pm$1.3 & 53.8$\pm$3.9 & 45.0$\pm$2.0 \\ \cmidrule(lr){2-2}
& \cmark & 42.8$\pm$2.1 & 36.2$\pm$0.6 & 51.5$\pm$2.2 & 40.7$\pm$1.3 & 53.8$\pm$3.9 & 45.0$\pm$2.0 \\ \cmidrule(lr){2-2}

JUMBOT~\cite{fatras2021jumbot} & \xmark & 36.8$\pm$0.2 & 33.6$\pm$1.3 & 50.5$\pm$0.6 & 35.6$\pm$3.0 & 63.7$\pm$3.0 & 44.0$\pm$1.6 \\
%\rowfont{\color{blue}} & \cmark & 42.7$\pm$0.1 & 38.3$\pm$1.2 & 49.7$\pm$4.2 & 41.5$\pm$0.5 & 62.2$\pm$1.2 & 46.9$\pm$1.4 \\ \cmidrule(lr){2-2}
 & \cmark & 42.7$\pm$0.1 & 38.3$\pm$1.2 & 49.7$\pm$4.2 & 41.5$\pm$0.5 & 62.2$\pm$1.2 & 46.9$\pm$1.4 \\ \cmidrule(lr){2-2}

\textbf{TimeMatch} & \xmark & \textbf{57.4$\pm$1.5} & \textbf{47.0$\pm$0.9} & \textbf{61.7$\pm$4.9} & \textbf{52.1$\pm$1.4} & \textbf{73.0$\pm$0.5} & \textbf{58.2$\pm$1.8} \\ 
\midrule
Target-trained & \xmark & 74.6$\pm$0.6 & 72.4$\pm$1.4 & 86.9$\pm$2.7 & 90.6$\pm$4.3 & 85.7$\pm$0.7 & 82.0$\pm$1.9 \\  
\bottomrule
\end{tabu}
\end{center}
\caption{Macro F1-score (\%) results on our dataset for unsupervised cross-region adaptation.
\blue{We consider five adaptation tasks across four Sentinel-2 tiles: DK1=32VNH (Denmark), FR1=30TXT (mid-west France), FR2=31TCJ (southern France), and AT1=33UVP (Austria).}
}
\label{tab:main_results_f1}
\end{table*}
\paragraph{TimeMatch}
We use the same training setup as the source-trained model but instead train for 20 epochs with a lower initial learning rate of $0.0001$. 
We define an epoch as 500 iterations to fix the frequency in which the temporal shift is  re-estimated.
We use maximum temporal shift $\Delta=60$ days, as we did not observe shifts greater than 2 months for our dataset in Europe.
We set the trade-off hyperparameter $\lambda=2.0$ in Equation~\ref{eq:timematch_loss}, EMA keep-rate $\alpha=0.9999$, and pseudo-label threshold $\tau = 0.9$. A sensitivity analysis of these hyperparameters is provided in Section~\ref{sec:sensitivity}.
For the FixMatch~\cite{sohn2020fixmatch} augmentations, we use the identity function for the weak $a(\cdot)$ in Equation~\ref{eq:pseudo_preds} and randomly sub-sample time steps for the strong $A(\cdot)$ in Equations~\ref{eq:loss_src_to_trg} and \ref{eq:loss_trg}. 
These are used for simplicity, and we leave the use of more advanced augmentations for SITS to future work.
At each iteration, we sample two mini-batches of size 128, one from the source and one from the target, in order to calculate the TimeMatch objective in Equation~\ref{eq:timematch_loss}.
We use a class-balanced mini-batch sampler for the source domain to ensure each source mini-batch contains roughly the same number of samples for each class. This reduces the class imbalance problem for the source domain for improved performance.
Additionally, we apply domain-specific batch normalization~\cite{li2016revisiting, chang2019domain, saito2019semi} by forwarding the source and target mini-batches separately instead of concatenated. This ensures the batch normalization~\cite{ioffe2015batch} statistics are calculated separately for each domain, for improved adaptation.

\paragraph{Competing Methods}
We re-implement the competitors MMD, DANN and CDAN+E following the domain adaptation library in~\cite{dalib}, and use the original source codes for \blue{ALDA~\cite{chen2020adversarial}} and JUMBOT~\cite{fatras2021jumbot}.
FixMatch~\cite{sohn2020fixmatch} follows our re-implementation for TimeMatch with an EMA teacher and the student as the final model.
All methods are initialized from a source-trained model. \blue{ShiftAug versions are initialized from the corresponding ShiftAug source-trained model, and we continue to use ShiftAug during training.}
As in TimeMatch, we train for 20 epochs and tune the hyper-parameters of these methods on the task DK1$\to$FR1. The full details can be found in our source code.

\section{Experimental Results}
\label{sec:experiments}
\subsection{Main Results}
Table~\ref{tab:main_results_f1} shows the performance obtained with our approach and the re-implemented baselines and competitors. We report the mean and standard deviation of macro F1 scores, calculated from the results of three runs with different dataset splits.

We observe that source-trained models transfer very poorly to new target regions, 
with an average F1-score of $39\%$ on target data. In comparison, target-trained models on the same classes achieve $82\%$ on average.
\blue{We observe that training shift-invariant models with ShiftAug improves domain generalization, leading to an increased average score of $49\%$. This greatly motivates addressing the temporal shift in UDA.}

\blue{Existing UDA methods, however, only slightly increase the performance of source-trained models, with the best result obtained by CDAN+E~\cite{long2017conditional} with $47\%$.
By incorporating our ShiftAug, we observe a performance boost across all evaluated methods,
indicating that existing methods are unable to implicitly handle the temporal shift.}

\blue{Our approach TimeMatch, where we explicitly estimate the temporal shift, outperforms all competing methods by $11\%$ on average and $5\%$ for their ShiftAug variants.
This shows that accounting for the temporal shift is a key component for the cross-region adaptation problem of SITS. Moreover, the shift-variant approach of TimeMatch outperforms the shift-invariance strategy. We hypothesize that training for shift-invariance may complicate crop classification, as the classification of certain crop types is shift-variant. For example, spring barley and winter barley develop similarly over time but shifted, as also discussed in Section~\ref{sec:introduction}.}

Comparing TimeMatch to the results of the target-trained model, we observe that our approach---without any target labels---recovers a significant part of the highest achievable performance if target labels were available, but we also find that there is room for improvement. \blue{From our results, we see that methods which explicitly account for the temporal shift, such as TimeMatch and the ShiftAug variants of competing methods, generally outperform methods which do not. We therefore believe that further improvements can be gained by considering stronger forms of temporal alignment than shifts, such as class-wise alignments or time warping. We leave this interesting direction to future work.}

\begin{figure}[ht]
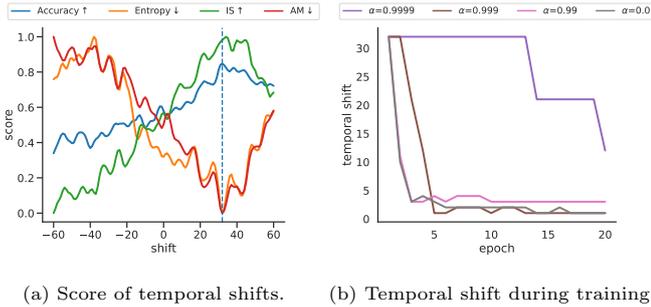

\centering
\begin{subfigure}[]{0.49\linewidth}
    \centering
    \includegraphics[width=1.0\linewidth, trim={0.0cm 0cm 0.0cm 0cm}]{\figfile{shift_score}}
    \caption{Score of temporal shifts.}
    \label{fig:shift_score}
\end{subfigure}
\begin{subfigure}[]{0.49\linewidth}
    \centering
    \includegraphics[width=1.0\linewidth, trim={0.0cm 0cm 0.0cm 0cm}]{\figfile{shift_epoch}}
    \caption{Temporal shift during training.}
    \label{fig:shift_epoch}
\end{subfigure}
\caption{(\subref{fig:shift_score}) Overall accuracy, entropy, IS, and AM scores of a source-trained model when applied to the target domain with different shifts. The dashed line indicate the most accurate shift.
(\subref{fig:shift_epoch}) The re-estimated temporal shifts of the teacher model during TimeMatch learning with different EMA decay rates.}
\label{fig:temporal_shift}
\end{figure}

\blue{Lastly, we highlight the results of the semi-supervised learning method FixMatch~\cite{sohn2020fixmatch}. 
This method is similar to TimeMatch, but without temporal shift estimation.
We observe that without ShiftAug, FixMatch obtains results worse than the source-trained model. 
This indicates semi-supervised learning cannot address the cross-region task alone. With ShiftAug, however, the results are greatly improved on average. Interestingly, the performance is worse without ShiftAug for all tasks \textit{except} FR1$\to$FR2. Here, the source and target regions are the most geographically close, and as result, the temporal shift is also closer to zero (see the top-left table in Figure~\ref{fig:tile_map}).
This indicates that ShiftAug (controlled by $\Delta$) is a trade-off between better long-range classification results and worse short-range results. In contrast, by estimating the temporal shift directly, TimeMatch does not have this issue and outperforms shift-invariance at both short and long distances.}

\subsection{Analysis of Temporal Shift Estimation}
In Figure~\ref{fig:shift_score}, we show the change in the overall accuracy of a source-trained model when applied to target data with different temporal shifts for DK1$\to$FR1. We also show the change in entropy, IS, and AM scores of the model.
We observe a significant increase in accuracy by temporally shifting the target data.
Calculating the statistics of entropy, IS, and AM from the predictions of the model works well as an unlabeled proxy to accuracy. 
We aim to estimate the shift with the highest accuracy (dashed blue line) for the highest quality pseudo-labels.
For the shown example, the minimum of both entropy and AM correspond to the best shift. However, we find AM to be the most consistent across different adaptation tasks.

In Figure~\ref{fig:shift_epoch}, we show the rate at which the estimated temporal shift for the teacher goes to zero in TimeMatch learning when training with different EMA decay rates.
When the shift changes, the previous estimate becomes sub-optimal for generating accurate pseudo-labels. We address this by re-estimating the temporal shift during training. We observe that low decay rates (\eg $0.99$) require the shift to be re-estimated after a few iterations, which is inefficient. In comparison, a decay rate of $0.9999$ allows us to only re-estimate the shift only once every epoch.

\begin{table}[ht]
\centering
\begin{tabular}{lc}
\toprule Ablation                                         & DK1$\to$FR1 \\
\midrule
No EMA ($\alpha=0.0$)                                     & 49.9$\pm$3.7    \\
No source temporal shift ($\delta^{s \rightarrow t} = 0$) & 51.9$\pm$1.9    \\ % write that shift will not move -- locked to source
No balanced batch sampler for source                      & 53.3$\pm$3.6    \\
IS instead of AM                                          & 56.3$\pm$2.6    \\
Entropy instead of AM                                     & 56.9$\pm$1.8    \\
No domain-specific batch norm.                            & 56.9$\pm$4.1    \\
% Cross Entropy loss             & 57.3$\pm$2.9       \\
\midrule
\textbf{TimeMatch}                                        & 57.4$\pm$1.5    \\
\bottomrule
\end{tabular}
\caption{Ablation study of TimeMatch components, sorted by increasing F1-score (\%).}
\label{tab:ablation}
\end{table}

The table in the upper left corner of Figure~\ref{fig:tile_map} shows the initial temporal shifts estimated by our method.
We find the estimated shifts are connected to the climatic differences between regions.
For example, the temporal shift ($\delta^{t \to s}$) from the warmer FR1 (mid-west France) to the colder DK1 (Denmark) is estimated as $32$ days. Due to the warmer climate, crops in FR1 mature earlier than in DK1, and a positive shift is required to align the former with the latter. In the other direction, the opposite is true, and indeed, we estimate a negative temporal shift of $-35$ days. Note that these are off by 3 days due to estimation variance. Here, the two source-trained models used to estimate the temporal shift in each direction are trained with two completely separate source region, yet their estimated shifts are still roughly inverses. This indicates that the temporal shift learned by these models is connected to the phenological properties of their respective source regions.

\begin{figure*}[ht]
\centering
\includegraphics[width=1.0\linewidth]{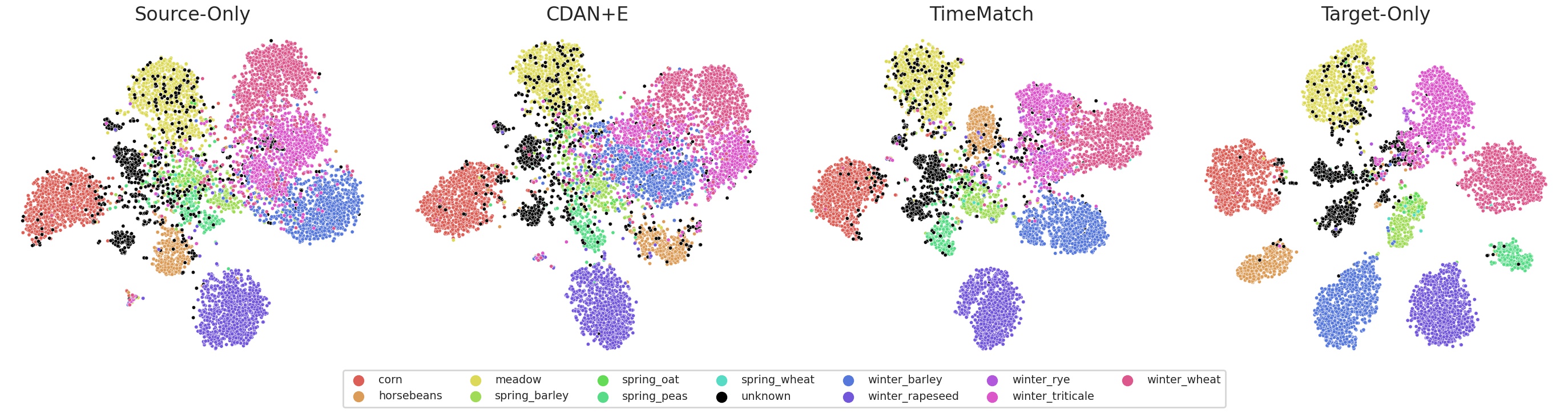}
\caption{Visualization with t-SNE~\cite{van2008visualizing} of target features for the DK1$\to$FR1 task. TimeMatch shows improved clustering of target features compared to existing approaches.}
\label{fig:tsne}
\end{figure*}

\begin{figure*}[ht]
    \centering
    \begin{subfigure}[b]{0.3\textwidth}
        \centering
        \frame{\includegraphics[width=0.7\textwidth]{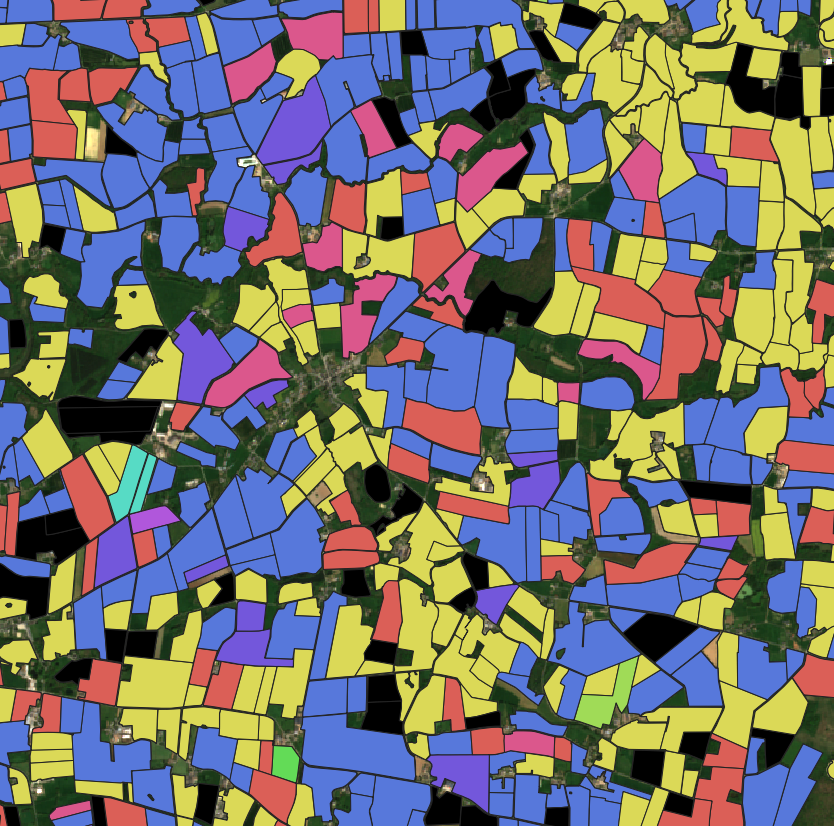}}
        \caption{Source-Only}
        \label{fig:pred1}
    \end{subfigure}
    \begin{subfigure}[b]{0.3\textwidth}
        \centering
        \frame{\includegraphics[width=0.7\textwidth]{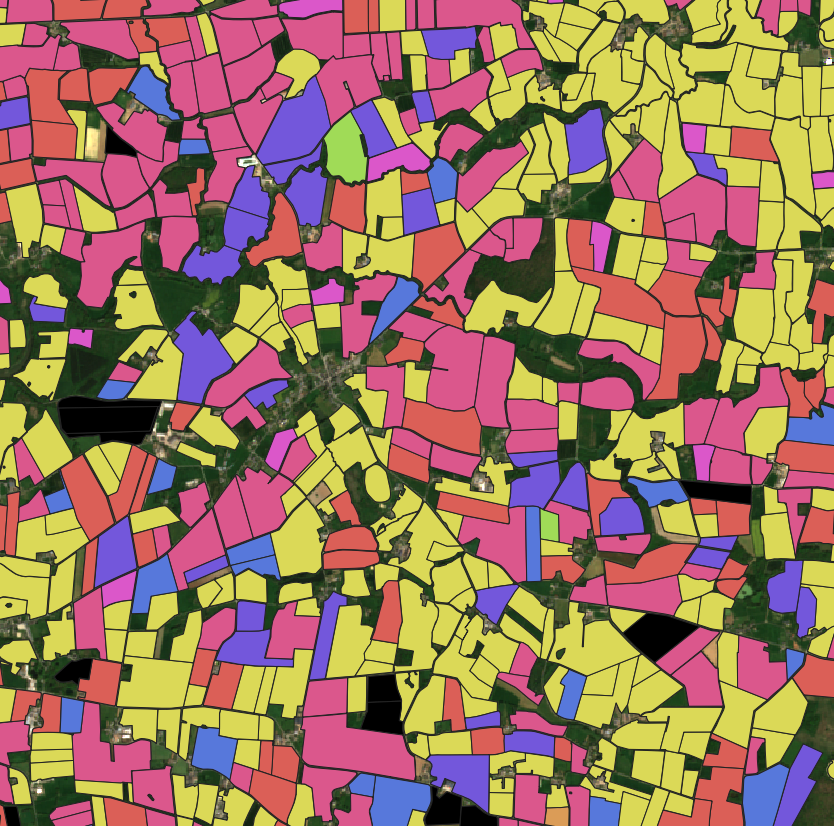}}
        \caption{TimeMatch}
        \label{fig:pred2}
    \end{subfigure}
    \begin{subfigure}[b]{0.3\textwidth}
        \centering
        \frame{\includegraphics[width=0.7\textwidth]{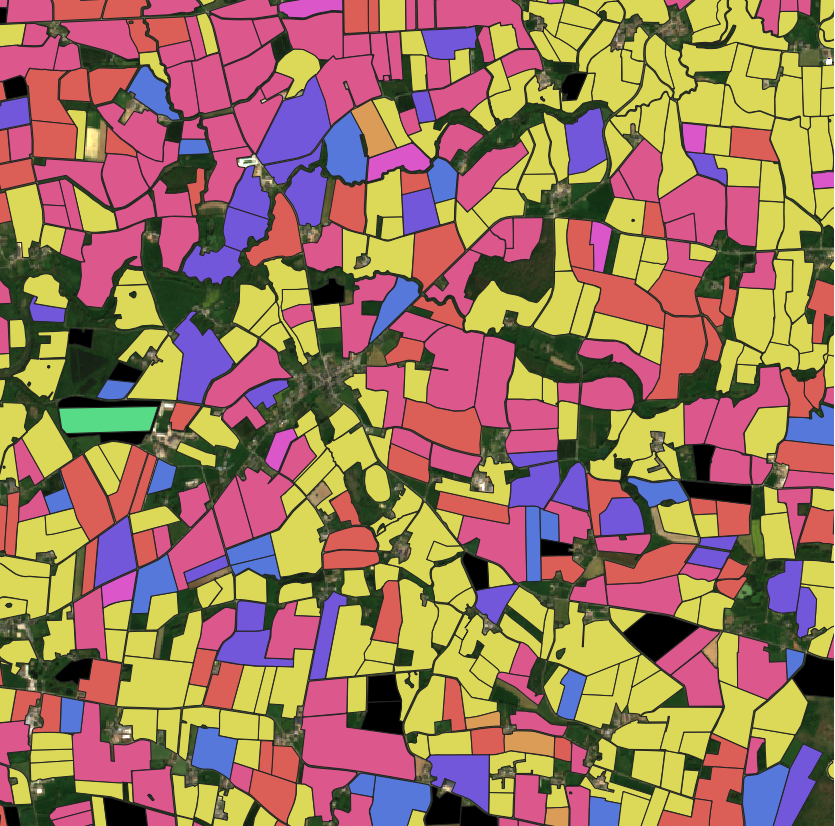}}
        \caption{Ground Truth}
        \label{fig:pred3}
    \end{subfigure}
    \caption{Parcel predictions for an example target area \green{(6 $\text{km}^2$)} from the DK1$\to$FR1 task, comparing (\subref{fig:pred1}) Source-Only, (\subref{fig:pred2}) TimeMatch, and (\subref{fig:pred3}) the corresponding ground truth. \blue{The figure shows the combination of multiple individual parcel predictions in the target region. The colors map to the classes in Figure~\ref{fig:tsne}.}}
\label{fig:preds}
\end{figure*}

\subsection{Ablation Study}
\label{sec:ablation}
To better understand how TimeMatch is able to obtain state-of-the-art results, we perform an ablation study on the different components for the task DK1$\to$FR1. We report the results in Table~\ref{tab:ablation}.
We first study the impact of the EMA training. Instead of EMA, we set the teacher as a direct copy of the student (No EMA). We observe that training without EMA introduces a significant drop in F1-score.
This shows that EMA is important to ensure high pseudo-label accuracy. 
Setting $\delta^{s \rightarrow t} = 0$ disables the temporal shift of the source domain, and the student is trained with datasets with different temporal shifts. We observe a significant decrease in F1-score as a result.
Disabling the balanced mini-batch sampler for the source domain also leads to a degradation of the performance. If the model is trained with class imbalanced source data, the teacher will make biased pseudo-labels for the samples from the target domain~\cite{he2009learning}.
This hinders the TimeMatch learning process, as pseudo-labels for infrequent classes in the source domain are less likely to be generated for the target.
By applying a balanced mini-batch sampler for the source, we address this problem by ensuring each source batch contains roughly the same number of samples for each category. Estimating the temporal shift with IS or entropy instead of AM results in a slight performance drop. Domain-specific batch normalization is simple to implement, as it just requires forwarding source and target batches separately instead of concatenated.
Disabling this component results in a small average performance loss with notably higher variance. 

\subsection{Sensitivity Analysis}
\label{sec:sensitivity}
\begin{figure}[ht]
    \centering
    \includegraphics[width=1.0\linewidth, trim={0.5cm, 0.5cm, 0.5cm, 0cm}]{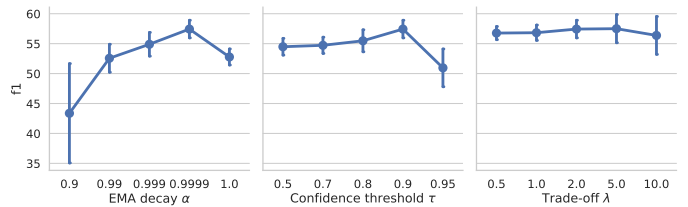}
    \caption{Sensitivity analysis of TimeMatch for the EMA decay rate, pseudo-label confidence threshold, and the trade-off in Eq.~\ref{eq:timematch_loss}. The error bars show standard deviation.} 
\label{fig:sens}
\end{figure}

Here we study the sensitivity of the TimeMatch hyperparameters. The results are shown in Figure~\ref{fig:sens}.
Higher values of $\alpha$ lead to better results, with a decay rate of $0.9999$ being the best. However, increasing it to $1.0$, so the teacher is not updated, results in a drop in F1-score, as the teacher cannot benefit from the knowledge learned by the student.
The confidence threshold $\epsilon$ controls the trade-off between the quality and quantity of pseudo-labels. A threshold of $0.9$ gives the best F1-score and further increasing the threshold to $0.95$ drops performance as a result of too few pseudo-labels, which particularly decreases performance for the less frequent classes. 
Finally, the trade-off parameter $\lambda$ controls the importance of the source domain loss $\mathcal{L}^{s \to t}$ with respect to the target domain loss $\mathcal{L}^{t}$.
We observe that this hyperparameter is less important than the other two, but setting $\lambda = 2.0$ gives the best results.

\subsection{Visual Analysis}
\blue{Finally, we visualize the ability of TimeMatch in learning discriminative features for the target domain}. In Figure~\ref{fig:tsne}, we visualize t-SNE~\cite{van2008visualizing} embeddings of target domain features from source-trained, CDAN+E (the best competitor on average), TimeMatch, and target-trained models on the task DK1$\to$FR1. The colors of the points represent their class (black is the unknown class).

With TimeMatch, the target features are better clustered into their respective classes compared CDAN+E, which does not result in much better feature separation than the source-trained model. The target-trained plot shows the best possible learned features when training with all available target labels. Even with labels, the classes are not perfectly separated, \eg for unknown/meadow or winter triticale/winter wheat.

Figure~\ref{fig:preds} shows example parcel predictions in a small area for the source-trained and TimeMatch models compared to the ground truth. 
The colors represent the same classes as before. We observe a large class confusion for the source-trained model, in particular between winter barley (blue) and winter wheat (dark pink), which are also not separated well in Figure~\ref{fig:tsne}. Without using any target labels, TimeMatch resolves this issue, resulting in clusters that better resemble the ground truth.

\section{Conclusion}
\label{sec:conclusion}
This paper presented TimeMatch, a novel cross-region adaptation method for SITS. Unlike previous methods that solely match the feature distributions across domains, TimeMatch explicitly captures the underlying temporal discrepancy of the data by estimating  the temporal shift between two regions. Through TimeMatch learning, we adapt a crop classifier trained in a labeled source region to an unlabeled target region. This is achieved by a learning algorithm that combines temporal shift estimation with \blue{self-training}, where target pseudo-labels are generated using the estimated temporal shift from target to source.
Lastly, we presented the TimeMatch dataset, a new large-scale cross-region UDA dataset with SITS from four different regions in Europe.
Evaluated on this dataset, TimeMatch outperforms all existing approaches by 11\% F1-score on average across five different adaptation tasks, setting a new state of the art in unsupervised cross-region adaptation.
\green{While this demonstrates that TimeMatch reaches strong results, there is still a gap with the performance obtained by fully supervised approaches. To overcome this limitation, we hypothesize that stronger temporal alignments, \eg class-wise alignments or time warping, could further improve the performance.}
\blue{Another possibility is to perform domain adaptation across both time and space, which in addition to the temporal aspect also brings new considerations, such as the change in parcel shapes over time and crop rotations.}
We hope our proposed method and released dataset will encourage the remote sensing community to consider the challenging cross-region adaptation problem and its temporal aspect. %In future work, we hope to see approaches that consider stronger forms of temporal alignment such as class-wise shifts or time warps.
    
\section{Acknowledgements}    
\blue{We would like to thank the anonymous reviewers for their insightful comments and constructive feedback.}
The work of Joachim Nyborg was funded by the \emph{Innovation Fund Denmark} under reference \emph{8053-00240}.

\bibliographystyle{elsarticle-num}
\bibliography{references}

\end{document}